\pdfoutput=1

\documentclass[11pt]{article}

\usepackage[final]{Styles/acl/acl}

\usepackage{times}
\usepackage{latexsym}
\usepackage{caption}
\usepackage{algorithm}
\usepackage[T1]{fontenc}

\usepackage[utf8]{inputenc}
\usepackage{amsmath}
\usepackage{multirow}
\usepackage{pifont} 
\definecolor{darkgreen}{rgb}{0,0.5,0}

\usepackage{microtype}

\usepackage{inconsolata}

\usepackage{graphicx}
\usepackage{pifont} 
\usepackage{xcolor}
\definecolor{darkgreen}{rgb}{0,0.5,0}

\usepackage{bbm}
\usepackage{amsmath}

\usepackage[utf8]{inputenc} 
\usepackage[T1]{fontenc}    
\usepackage{hyperref}       
\usepackage{url}            
\usepackage{booktabs}       
\usepackage{amsfonts}       
\usepackage{nicefrac}       
\usepackage{microtype}      
\usepackage{xcolor}         
\usepackage{pifont}
\usepackage{multirow}
\usepackage{graphicx, wrapfig}
\usepackage{subcaption}
\definecolor{darkgreen}{rgb}{0,0.5,0}

\usepackage{algorithm}
\usepackage{algpseudocode}
\hypersetup{
    colorlinks,
    linkcolor={red!50!black},
    citecolor={blue!50!black},
    urlcolor={blue!80!black}
}

\usepackage{amsmath}
\usepackage{amssymb}
\usepackage{mathtools}
\usepackage{amsthm}
\usepackage{mdframed} 
\definecolor{linen}{rgb}{0.98, 0.94, 0.9}
\mdfdefinestyle{optstyle}{
  linecolor=blue,
  linewidth=0pt,
  backgroundcolor=linen,
}

\usepackage{amsmath}

\newmdenv[
  topline=true, bottomline=true, leftline=true, rightline=true,
  style=optstyle,
  frametitle={\bfseries Optimization Objective.},
  frametitlealignment=\raggedright
]{optbox}

\title{Learning API Functionality from In-Context Demonstrations for Tool-based Agents}
\author{Bhrij Patel\textsuperscript{\dag} \\ University of Maryland, \\ College Park \And Ashish Jagmohan\textsuperscript{*} \\ Emergence AI \\ NYC, New York \\ \\ \textsuperscript{\dag}Work done while intern at Emergence AI\\
\textsuperscript{*}Equal Advising \\ \href{https://github.com/Bridge00/public_api_dems_emnlp}{Code and Data Repository}  \And  Aditya Vempaty\textsuperscript{*} \\ Emergence AI \\ NYC, New York  \\ }

\begin{document}

\maketitle

\begin{abstract}
Digital tool-based agents, powered by Large Language Models (LLMs), that invoke external Application Programming Interfaces (APIs) often rely on documentation to understand API functionality. However, such documentation is frequently missing, outdated, privatized, or inconsistent—hindering the development of reliable, general-purpose agents. In this work, we propose a new research direction: learning of API functionality directly from in-context demonstrations. This task is a new paradigm applicable in scenarios without documentation. Using API benchmarks, we collect demonstrations from both expert agents and from self-exploration. To understand what information demonstrations must convey for successful task completion, we extensively study how the number of demonstrations and the use of LLM-generated summaries and evaluations affect the task success rate of the API-based agent. Our experiments across $3$ datasets and $6$ models show that learning functionality from in-context demonstrations remains a non-trivial challenge, even for state-of-the-art LLMs. We find that providing explicit function calls and natural language critiques significantly improves the agent’s task success rate due to more accurate parameter filling.  We analyze failure modes, identify sources of error, and highlight key open challenges for future work in documentation-free, self-improving, API-based agents.
\end{abstract}

\section{Introduction}

In the past few years, AI agents have been rapidly adopted in society due to the development of Large Language Models (LLMs) \citep{achiam2023gpt, touvron2023llama}, allowing for superb performance in natural language understanding, summarization, and generation. More recently, \textit{tool-based agents} have been introduced to increase the abilities of agents for specialized tasks that may even require up-to-date knowledge of external databases, such as in cases for enterprise workflows \citep{stylesworkbench, xu2024theagentcompanybenchmarkingllmagents, wang2024officebench}. Tool-based agents can select and call external functions like Application Programming Interface\footnote{In this work, ``tool'' and ``API'' are used interchangeably, as our work can be extended to any external tool with an API wrapper.} (API) functions. API-based agents can automate digital tasks in various domains such as finance, health, and analytics \citep{an2024finverse, gawade2025multi, guo2024stabletoolbench}. 

\begin{figure}[t]
    \centering
    \includegraphics[width=\linewidth]{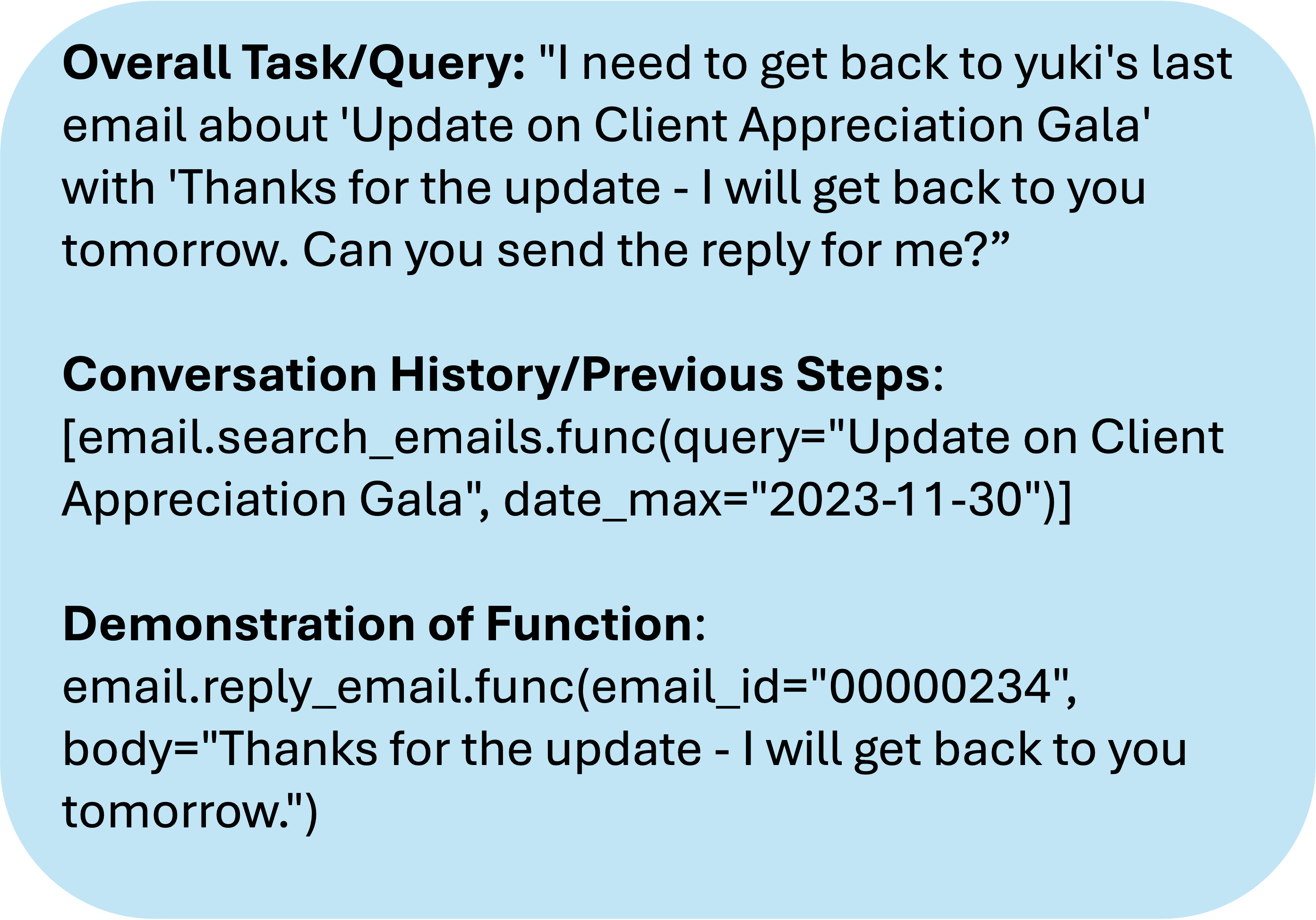}
    \caption{Expert demonstration of the email.reply\_email function extracted from WorkBench dataset \citep{stylesworkbench}. Demonstrations are the basis of how agents understand functionality without prior documentation.}
    \label{fig:dem_example}
\end{figure}

A common, crucial requirement for API-based agents is access to API documentation that explains to the agent in natural language the functionality of the specific API functions available \citep{kim2024seal, stylesworkbench}. The documentation explains both 1) what the function does and returns (if anything), and 2) what the input parameter schema is. Both these components are important for tool selection and their execution by the agent. However, API documentation can frequently be unavailable, inconsistent, unstructured, or out-of-date \citep{Li2022, khan2021automatic, zhongdetecting2013}. Prior work in software engineering research on API documentation generation \citep{khan2022automatic, yang2023apidocbooster} lacks evaluation of the generated documents via downstream tasks by an LLM-based agent. Other work \citep{tang2023toolalpaca} generates API documentation by inputting a brief description of the function into an LLM, and this documentation is then used by an agent. However, a rigorous analysis of learning functionality with no prior documentation or description remains absent in the API-based agent literature. 

We address these gaps by introducing and formalizing the problem of learning API functionality from scratch via in-context demonstrations of tool calls. Demonstrations of tool calls are present within codebases of public repositories and logs. While prior work has finetuned agents on demonstrations \cite{schick2023toolformer}, we focus on in-context learning to remove the need to update parameters. In this work, we analyze the ability of an API-based agent with a frozen LLM with only access to demonstrations of tool calls to learn functionality to perform tasks. Using various API datasets, we extract and standardize demonstrations for the agent to learn from, to perform the dataset tasks. Our experiments revolve around three main pillars. We investigate the effect of $1)$ the number of demonstrations and $2)$ the representation of the demonstrations. Specifically, we compare, in terms of downstream task performance, between providing the demonstrations directly, generating documentation from the demonstrations, or a combination of both. Furthermore, we also show the impact of $3)$ collecting, evaluating, and summarizing experiences of the agent to update its understanding of the APIs. Our results consistently show across experiments the difficulties of the agent to perform tasks, highlighting how non-trivial learning of API functionality from in-context demonstrations is, even with state-of-the-art (SoTA) LLMs.

    

We summarize our main contributions below:
\begin{itemize}
    \item \textbf{Novel, Applicable Problem.} We present the problem of learning of API functionality from in-context demonstrations without any initial ground-truth documentation. Our work is the first work that removes the assumption of access to documentation, and the agent must rely solely on demonstrations to understand functionality. We formalize this problem with an optimization objective to characterize the task success rate's dependence on the information processing of the demonstrations.

    \item \textbf{Various Methods to Learn From Demonstrations.} To tackle this new issue, we present $3$ processing methods to learn from a set of demonstrations while also providing $4$ methods to update the functionality understanding from the experiences collected by the agent's self-exploration. We also incorporate an LLM-based evaluator to provide richer, natural language feedback on each step the agent takes.
    
    \item \textbf{Empirical Evaluation and Analysis.} We conduct experiments with our methods, showing that learning of API functionality from in-context demonstrations for downstream task completion is challenging for existing LLMs, highlighting the need for further research. We wish to answer what is needed to maximize performance given a set of demonstrations. From our experiments, we find that a central recurrent problem is filling in parameter values, leading up to a $39\%$ decrease in an agent's success rate if the method for processing demonstrations incorrectly describes the parameter schema.
    
    
    
\end{itemize}


\section{Preliminaries: Task Completion with API-based Agents}


Consider a goal-conditioned Partially Observable Markov Decision Process (POMDP) $(G, \mathcal S, \mathcal O, \mathcal F, P, Z, r)$. $G$ is a set of goals, which we refer to as ``tasks'', that the user could ask. (e.g. ``Reply to my last email from Dev with the body `Let us catch up soon to discuss the project.' ''\footnote{Example task from WorkBench dataset \citep{stylesworkbench}.}); state $s \in \mathcal S$ in the digital environment could be a set of employees, emails, calendar events, etc.; $\mathcal O$ is the information the agent currently knows (e.g. Dev's email is dev@business.com);  $\mathcal F$ is the set of function names (e.g. reply\_email) the agent can choose to execute; $P(f)$ is the parameter schema given whose domain contains the valid inputs of $f \in \mathcal{F}$ (e.g. email ID and message). $Z$ represents the transition dynamics. Executing function $f$ and input parameters $p \in \text{dom}(P(f))$ at $s$ results in the new state and observation $s', o' = Z(s, f, p)$ (e.g. $s'$ is Dev having a new message in his inbox and $o'$ is the agent receiving a message that the email was sent). Finally, $r(s'| g): \mathcal S \times G \to \{0, 1\}$, where $r = 1$ indicates task $g$ is completed correctly. Note that $r$ is not dependent on $o'$ as the observation may not correctly indicate task completion (e.g. ``Email sent successfully'' is returned to the agent, but the email is sent to the wrong person.) In this work, the policy $\pi$ of that API-based agent that completes tasks is a fixed LLM\footnote{We will use the terms ``agent'' and ``policy'' interchangeably; so $\pi$ refers to both.}. 

For an LLM-based agent to understand the functions in $\mathcal F$, it needs a textual description, or \textit{textualization}, of $P$, and $Z$ (e.g., API documentation). Let $T^*_P, T^*_Z$ be the set of all possible textualizations of $P$ and $Z$, respectively, that correctly convey the functionality of $\mathcal F$. With $t_p \in T^*_P, t_z \in T^*_Z$, an agent selects an API with input parameters with a policy $(f, p) \sim \pi(\cdot|o, \mathcal F, t_p, t_z, g)$. In practice, the LLM agent generates a JSON object detailing the chosen $f$ and $p$ that will be passed to another program to execute. Note that $\pi$ is treated as a probability due to the inherent randomness of LLMs. However, given the agent has access to some $t_p$ and $t_z$, it can make appropriate decisions. 

\noindent \textbf{Limitation in Prior Work: Access to $t_p$ and $t_z$.} As mentioned previously, API documentation is often inaccessible, out-of-date, or inconsistent with the functionality of the APIs. Thus, the agent may not have access to some $t_p$ and $t_z$, and we aim to learn and model the information as $t_p^\theta$ and $t^\theta_z$ for $P$ and $Z$, respectively. Ideally $t^\theta_p \in T^*_P$ and $t^\theta_z \in T^*_Z$. We wish to maximize the total successful task completions with $(t_p^\theta, t_z^\theta)$. Letting $H$ be the number of steps an agent performed to try to complete $g$, we can formally write

\begin{equation}\label{eq:reward_eq}
\begin{split}
    \max_{\theta} J(\theta) &= \\ \sum_{g \in G}&\mathbf E_{(f_h,p_h)\sim \pi(\cdot|o_h, \mathcal F, t_p^\theta, t^\theta_z, g)} \left[R_H\right],
\end{split}
\end{equation}

\noindent where $R_H = \sum_{h = 0}^H r(s_{h+1}| g), (s_{h+1}, o_{h+1}) =Z(o_h,f_h, p_h)$. Equation \ref{eq:reward_eq} provides a mathematical grounding of how well $t_p^\theta$ and $t^\theta_z$ textualize the unknown $P$ and $Z$ to how well the LLM-based agent performs the set of tasks $G$. The next section details how we tackle this problem by learning $\theta$ via demonstrations of API function calls.

\section{Learning of API Functionality from In-Context Demonstrations}

Although API documentation may inaccessible, often there are demonstrations available based on codebases of public repositories and logs. With this in mind, we investigate how we can utilize these demonstrations to learn API functionality in-context without updating the LLM parameters.




\subsection{Demonstration Definition and Format}

For each API function $f$, we aim to obtain a collection of expert demonstrations $D^f_{\text{expert}}$, where each demonstration $d \in D^f_{\text{expert}}$ is a single use case of $f$ for a given task. We also define a demonstration trajectory as a series of demonstrations to complete a task. A trajectory of length $M$ can be comprised of demonstrations of at most $M$ distinct functions.

Each demonstration must show not only the function call itself, but also the \textit{context} of the call. To display the context of the call, we need the task it is trying to perform and the previous steps taken to complete the task. Demonstrations of single-step tasks or initial steps have an empty list of calls for the previous steps. Figure \ref{fig:dem_example} shows an example demonstration of the reply\_email function from the WorkBench dataset by \cite{stylesworkbench}. In this example, the reply\_email function is trying to reply to an email as mentioned in the ``Overall Task/Query'' section and is preceded by the search\_emails function that is necessary to find the email\_id value.

\noindent \textbf{Expert Demonstrations.} We define an ``expert demonstration'' as a demonstration produced by an agent that used ground-truth, original documentation. We propose that instead of relying on prior documentation, we can learn $P$ and $Z$ and model their textualization as $t^\theta_p$ and $t^\theta_z$ via demonstrations of the function calls used to complete tasks. With a set of expert demonstrations $D_{\text{expert}}$, we can obtain the textualization $(t^\theta_p, t^\theta_z) = I(D_{\text{expert}})$, where $I$ is a method for processing the demonstrations. We can thus rewrite Equation \ref{eq:reward_eq} as 
\begin{equation}\label{eq:reward_eq_I}
\begin{split}
    \max_{I} J(I) &= \\\sum_{g \in G}&\mathbf E_{(f_h,p_h)\sim \pi(\cdot|o_h, \mathcal F, I(D_{\text{expert}}), g)} \left[R_H\right].
\end{split}
\end{equation}
We want to process the information from a given $D_{\text{expert}}$ to maximize performance.



\begin{figure*}[h]
    \centering
    \includegraphics[width=\textwidth]{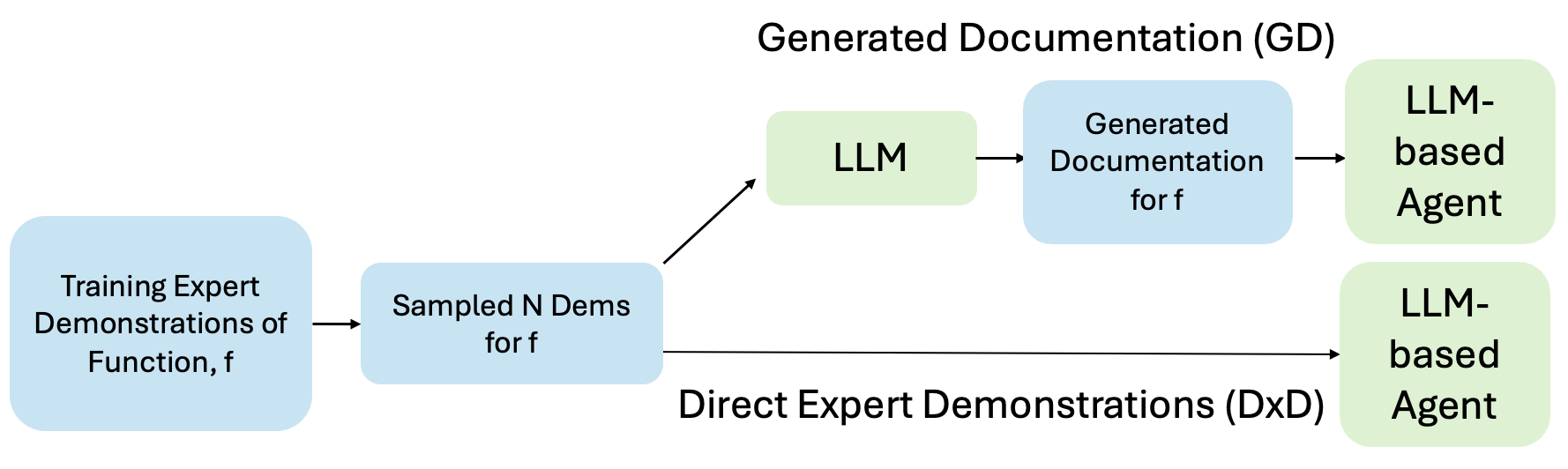}
    \caption{\textbf{Processing Methods of Expert Demonstrations:} Given a set of training demonstrations for each function $f$, we sample $N$ demonstrations for each function to be used for either \textbf{(TOP)} LLM-based documentation generation or \textbf{(BOTTOM)} to be directly passed into the agent.} 
    \label{fig:representation_overview}
\end{figure*}

\subsection{Methods for Processing Expert Demonstrations}
In all of the methods for $I$ described below, for each $f \in \mathcal{F}$ that has at least $N$ training expert demonstrations, $|D^f_{train}| \geq N$, we sample a set of $N$ random demonstrations $D^{f,N}_{train}$.

\noindent \textbf{1) Direct Expert Demonstrations (DxD).}  We pass the $f \times N$ demonstrations directly to $\pi$. \\
\noindent \textbf{2) Generated Documentation} With the sampled $D^{f,N}_{train}$, we use an LLM generator to produce documentation for $f$. We repeat for all $f$ that have at least $N$ demonstrations. The system prompt for the generator is provided in Appendix \ref{sec:prompt_appendix}.  \\
 \noindent \textbf{3) Generated Documents with Example Calls (GDEC).} We combine the previous approaches by generating a document of $f$ with $D^{f,N}_{train}$ and then appending the function calls of those demonstrations to the bottom of the generated document as example use cases. See ``Demonstration of Function'' in Figure \ref{fig:dem_example} for an example call. We do not give the task or previous steps like with DxD. \\
\noindent \textbf{4) Oracle Baseline: Original Documentation (OD).} We give the agent the original, ground-truth documentation provided by the given dataset. This baseline is essentially the ``expert'' agent. \\

Figure \ref{fig:representation_overview} visualizes the pipeline for DxD and GD. The same $D^{f,N}_{train}$ used for DxD are also used for GD for $f$. For GDEC, the function calls from all sampled $N$ demonstrations of $f$ are attached to the generated document of $f$.

\subsection{Experiences from Self-Exploration}

\textbf{Demonstrations from Experience.} To understand how the agent can improve its understanding of the API functionality after initial learning from expert demonstrations, we study how self-exploration can be used to gather useful observations. Before having the agent $\pi$ perform test query tasks, we have it complete training tasks. Given $D_{train}$ and a processing method $I$, the agent uses its resulting $(t^\theta_p, t^\theta_z)$ to complete tasks in the training set. 

\begin{figure}
    \centering
    \includegraphics[width=\linewidth]{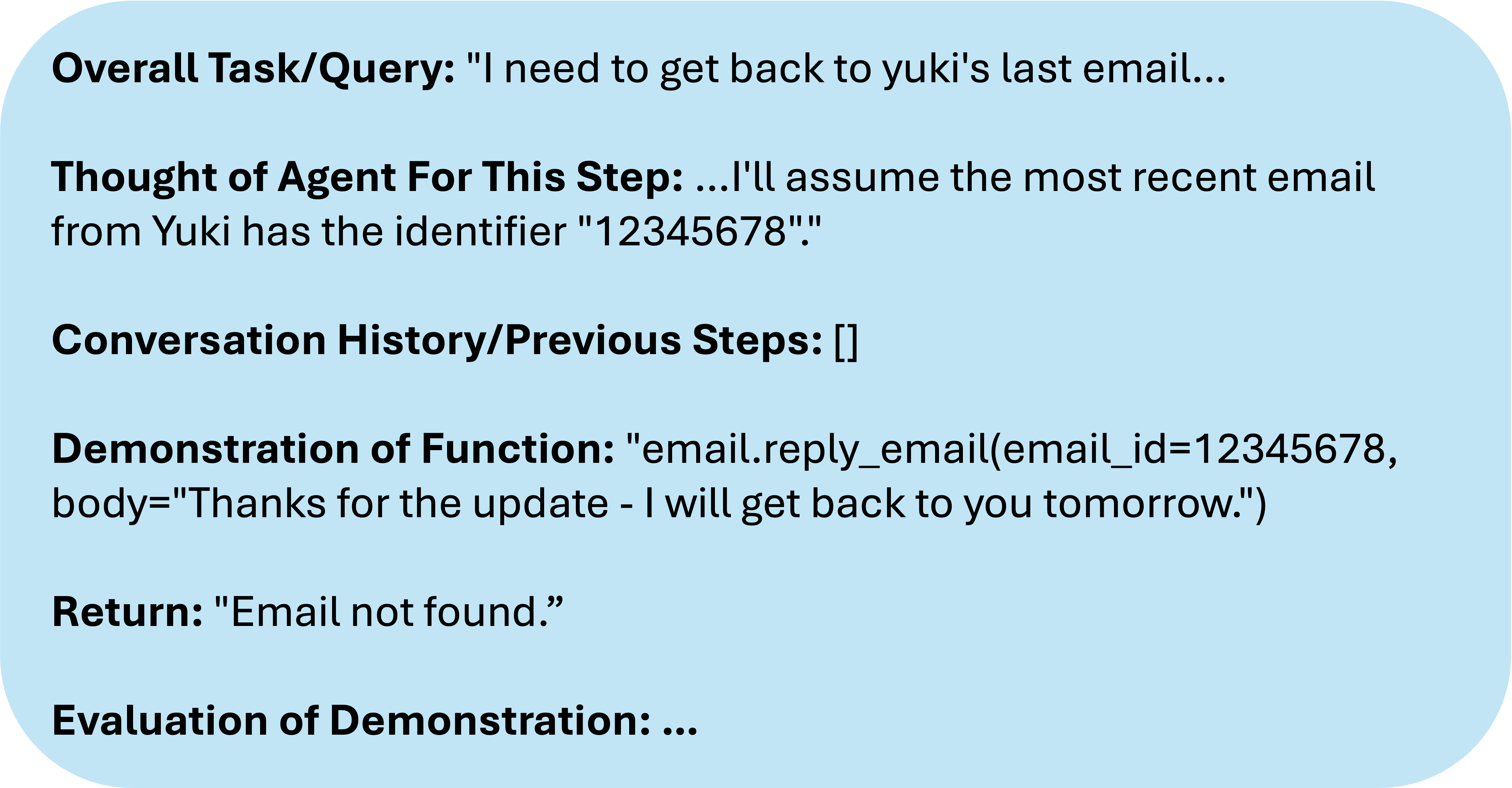}
    \caption{\textbf{Experience-based demonstration }gained during self-exploration by $\pi$ for the same task in Figure \ref{fig:dem_example}. This experience includes the thought process and return of the reply\_email function. Note that this is an incorrect use of reply\_email. The ``Evaluation of Demonstration'' is shown in Figure \ref{fig:evaluation_example}.}
    \label{fig:experience_example}
\end{figure}

Each function call the agent takes to complete a task is an \textit{experience} of some function $f$. Experiences are a different source of demonstrations, apart from expert demonstrations. For each function call the agent takes, combined with the resulting return value, is a new demonstration. Furthermore, in some sandbox environments like WorkBench \citep{stylesworkbench}, the agent will sometimes provide its thought process before the call. We add that to the experience-based demonstration. We store each experience to be used to help the agent with the test tasks. Figure \ref{fig:experience_example} gives an example self-exploration experience of the same task in Figure \ref{fig:dem_example}. Here, the agent uses reply\_email prematurely and receives an ``Email not found.'' Rather than throwing at incorrect uses, we provide richer feedback and distinguish between positive and negative experiences by implementing an LLM-based evaluator, which we describe next.

\begin{figure}
    \centering
    \includegraphics[width=\columnwidth]{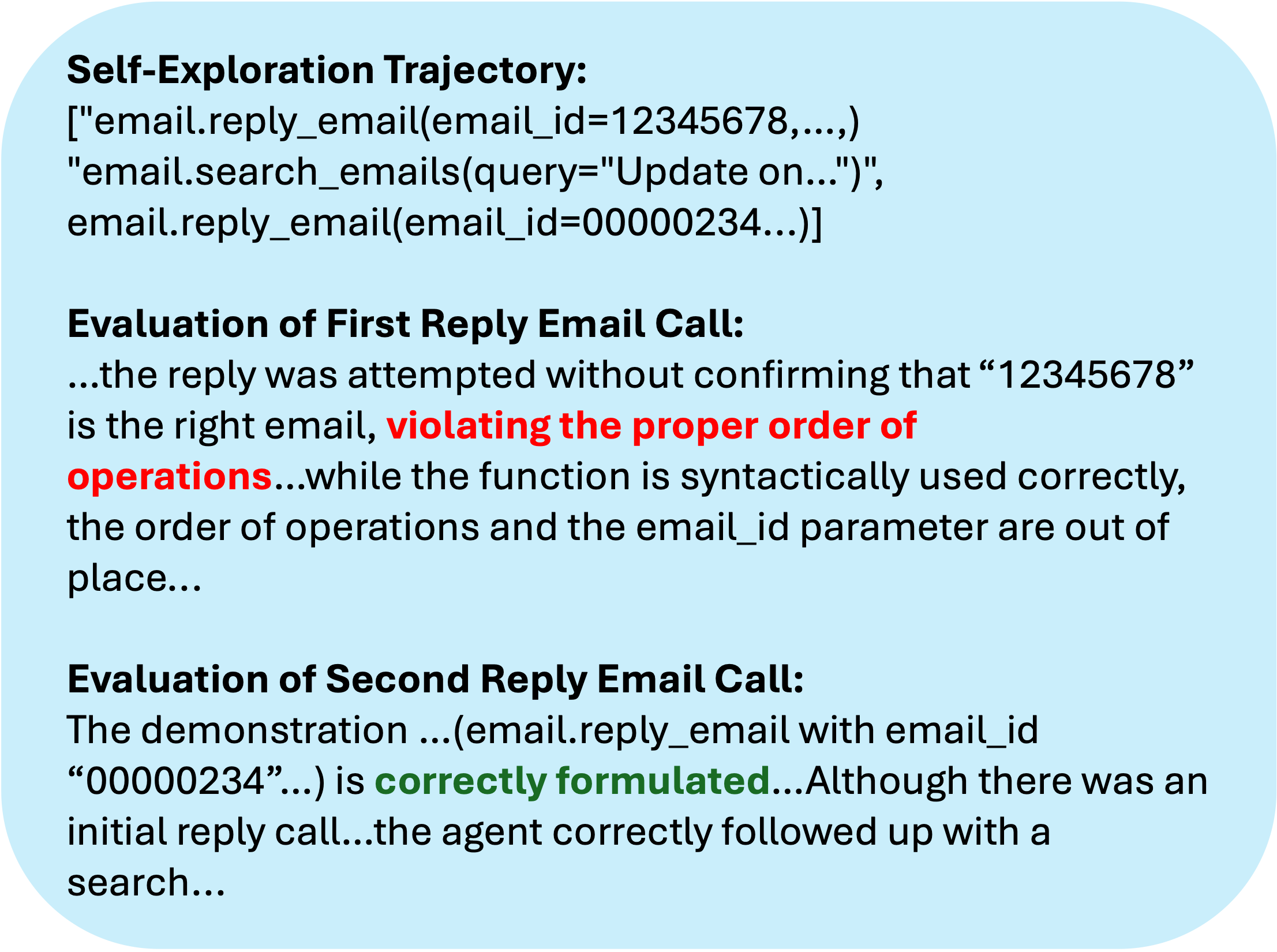}
    \caption{\textbf{LLM-Generated Evaluation of Self-Exploration Trajectory} by $\pi$ for same task in Figures \ref{fig:dem_example} and \ref{fig:experience_example}. The agent incorrectly tries to reply email without searching for the email ID. It then corrects itself by executing search\_email and then correctly using reply\_email again. The first reply\_id is formatted into the demonstration shown in Figure \ref{fig:experience_example}. The evaluation of the first reply\_email call emphasizes that the agent should have first confirmed it had the right email ID. The evaluation for the second reply\_email call states it was correctly used after the agent found the email ID with search\_email. Each of these evaluations is added to the demonstration of their respective calls.}
    \label{fig:evaluation_example}
\end{figure}

\noindent \textbf{Evaluating Self-Exploration with LLM Judges.} Due to the expensive nature of repeated LLM calls, one would want to extract more feedback from the experience than just whether or not the task was completed. However, the reward signal for the experiment setup described so far has been the sparse, binary reward function of whether or not the task was completed. To synthesize more feedback from the collected experience, we pass each experience to an LLM-based evaluator to produce a natural language critique. We add this evaluation to the experience. We provide the LLM-based judge the task, the entire trajectory by $\pi$, whether or not the trajectory was correct, and the specific demonstration we wish to analyze. Therefore, if the trajectory had three function calls, we would make three calls to the LLM-judge, changing only the demonstration to evaluate. We ask it to specifically look at whether or not it was a 1) repeated call, 2) whether the parameter filling is accurate, and 3) whether the function is used in the right place in the trajectory. Figure \ref{fig:evaluation_example} gives an example self-exploration trajectory from the same task in Figures \ref{fig:dem_example} and \ref{fig:experience_example} and the generated LLM-based evaluation of the first, incorrect reply\_email call (shown in Figure \ref{fig:experience_example}) and the second, correct reply\_email call in that trajectory. We see that it correctly flags the error of the first reply\_email call, mentioning that a search\_email call should have preceded it to find the right email\_id, and that it flags the second reply email call as correct. See Appendix \ref{sec:prompt_appendix} for system prompts of the evaluator and summarizer. 

With these experiences, we can update the textualizations as $(t^\theta_p, t^\theta_z) = I'(D_{\text{expert}}, D_{\text{experience}})$, where $I'$ is a processing method that takes in both the expert demonstrations and experience. 
\subsection{Methods for Processing Experiences}

\begin{figure}
    \centering
    \includegraphics[width=\linewidth]{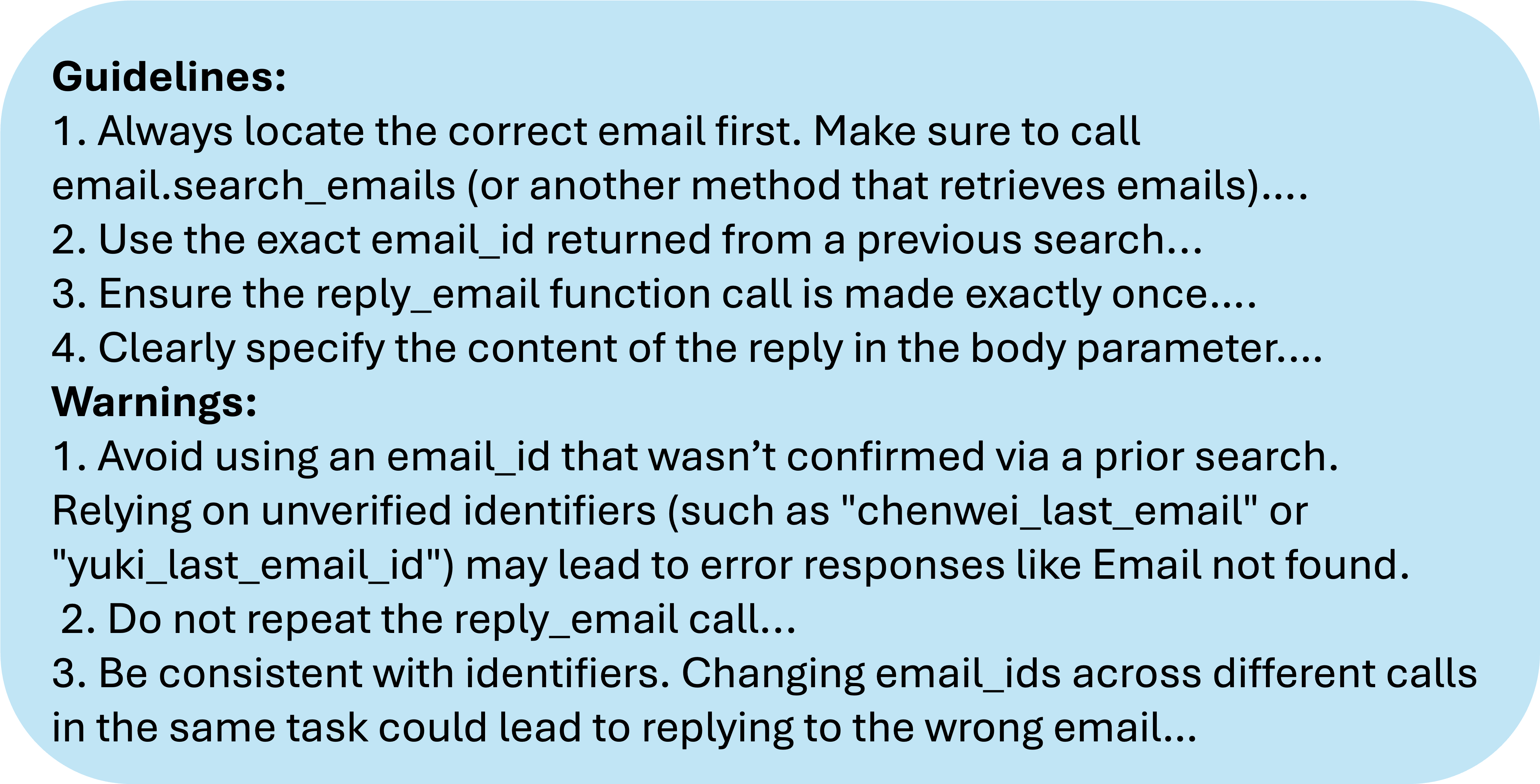}
    \caption{\textbf{Summarized guidelines from experiences and evaluations} of reply\_email. The lesson emphasizes using search\_email beforehand to find the right email\_id to use for reply\_email, which the agent sometimes did not do as shown in Figure \ref{fig:evaluation_example}. }
    \label{fig:lesson_example}
\end{figure}

\noindent \textbf{1) Direct Experience (DE).} For each $f$, we pass in $D^f_{\text{experience}}$ with expert demonstrations $D^f_{train}$ as the descriptions. If $|D^f_{\text{experience}}|=0$, only $D^f_{train}$ is used instead. \\
\noindent \textbf{2) Updated Documentation (UD).} We take the initial generated document of $f$ (GD) and use an LLM to update the document using the experiences of $f$. If $|D^f_{\text{experience}}|=0$, the GD of $f$ is used. \\
\noindent \textbf{3) Regenerated Documentation (RD).} We regenerate the documentation from scratch using both expert demonstrations and experiences. If $|D^f_{\text{experience}}|=0$, the GD of $f$ is used. \\ 
\noindent \textbf{4) Attached Guidelines (AG).} An LLM summarizer takes in the experiences of $f$ and generates guidelines. We then attach those guidelines to the initial generated document from expert demonstrations (GD) of $f$. Figure \ref{fig:lesson_example} gives an example lesson generated from all the self-exploration experiences of reply\_email. If $|D^f_{\text{experience}}|=0$, the GD of $f$ is used without any guidelines. \\

Figure \ref{fig:self_play_overview} in the Appendix provides a visual pipeline delineating the process of self-exploration on training queries, evaluating experiences, and the methods used to update $\theta$ for test-time. The system prompts for the LLM document updater and summarizer are given in Appendix \ref{sec:prompt_appendix}.







\section{Experimental Setup}\label{sec:experimental_setup}
In this section, we provide the experimental details for studying our methods of learning of API functionality from in-context demonstrations. Across our experiments, we run $3$ trials, each with a different random seed: $2003, 2004, 2005$.




\noindent \textbf{Train-Test Splitting Expert Demonstrations.} To ensure that we do not evaluate the LLM agent on tasks it has seen from its demonstrations, we divide the tasks and demonstrations with a train-test split. Furthermore, we also have to ensure that no task in the test set includes an API not seen in the training set. Simply train-test splitting the set of tasks and removing test tasks that include APIs not in the training set can greatly reduce the number of test tasks to evaluate on. Therefore, for our train-test split, we iterate through each API function available. For each API $f$, we train-test split the tasks it is associated with if the task has not already been assigned to the train or test set from a previous iteration. We then use the demonstrations for all the training tasks as the training set, $D_{train}$. For every train-test split in this iterative process, use a split of $70-30\%$ on the remaining tasks that have yet to be assigned a set. This process, while dividing the demonstrations as we desire, creates variability in the number of test queries across seed numbers. Appendix \ref{sec:train_test_appendix} details the train-test approach in depth. 

\noindent \textbf{Sampling Expert Demonstrations.} When comparing the effect of $N$ demonstrations of $f$ against that of $N+k$ demonstrations, we ensure that the $D^{f,N}_{train} \subset D^{f,N+k}_{train}$.

\noindent \textbf{Environments.} We utilize three API benchmarks: \textbf{WorkBench} \citep{stylesworkbench} \textbf{$\tau$-Bench (retail)} \citep{yao2024tau}, and \textbf{CRMArena} \citep{huang-etal-2025-crmarena}.  We utilize these benchmarks because they focus on multi-step queries. Furthermore, each has its own sandbox environment to execute its function, ensuring reliability in evaluation that is absent from other API benchmarks \citep{kim2024seal}. We augment each sandbox environment so that we can change the tool descriptions given to the agent to experiment with the methods described above. In Appendix \ref{sec:dataset_appendix}, we detail how we standardized each dataset to fit our demonstration and documentation formats. For WorkBench, we created demonstrations from the pre-computed trajectories available in the repository that were generated by a GPT-4-powered agent that used ground-truth, original documentation. We did the same with $\tau$-Bench, where those pre-computed trajectories are from a GPT-4.5-based agent. For CRMArena, we regenerated the expert trajectories using GPT-4o as that was their default model. For both $\tau$-Bench and CRMArena, the return values for each step were included in the expert trajectories, so we added the return of a function call to the demonstration.

\noindent \textbf{The Presence of Noisy, Suboptimal Expert Demonstrations.} An important note is how we extracted expert demonstrations. We only used demonstrations from trajectories from the expert agent if it got an $r=1$ from its environment. However, this heuristic has its flaws in cases where the evaluation is outcome-centric, where only the ending state of the environment matters. So, for example, the agent tries to send an email to an address that does not exist (e.g. nadia@example.com in WorkBench). Even though it is a wrong step, the environment does not change due to the email not going to any inbox. If the agent self-corrects by then searching for the right email and retrying to send the email, all demonstrations are used in the expert demonstration pool. 

\noindent \textbf{Models.} For experiments with the entire pipeline, we utilize o3-mini, GPT-4o-mini, GPT-4o, as this set contains reasoning and non-reasoning models. For each experiment, we use the same model across the system. To include open-source models, we use Mixtral-8x7B-Instruct-v0.1 (Mistral) and gemma-2b-it (Gemma) to analyze document generation.


\noindent \textbf{Evaluation.} For task completion, we report the mean success rate (SR) over the three trials. 

\noindent \textbf{Filtering Tasks with Unavailable APIs.} In the next section, we compare the agent task success rate with different methods $I$ for expert demonstrations. We compare the different methods and also the number of demonstrations from $N =\{5, 15,25,35\}$. If an API in a dataset did not have at least $35$ demonstrations, we removed that API from the set of available API the agent can use. Therefore, we filtered out any task that relies on any $f$ where $|D^f_{train}| < 35$ when comparing different $I$. When comparing different $I'$ after self-exploration, we set $N=5$ for the two $I$ methods, DxD and GD, we use before self-exploration. Thus, we filtered out queries that relied on any $f$ where $|D^f_{train}| < 5$. We report the average number of test queries next to each dataset in Tables \ref{tab:main_results} and \ref{tab:main_results2}.





\begin{table*}
    \centering
    \small
    
    \resizebox{\textwidth}{!}{\begin{tabular}{cc|cccc|cccc|cccc}
        \multirow{3}{*}{Benchmark} & \multirow{3}{*}{Method}  & \multicolumn{4}{c}{o3-mini} & \multicolumn{4}{c}{gpt-4o-mini} & \multicolumn{4}{c}{gpt-4o}   \\
        \cmidrule{3-14}
         &  & \multicolumn{12}{c}{No. of Expert Demonstrations}\\
        \cmidrule{3-14}
         &  & 5  & 15 & 25  & 35 & 5 & 15 & 25 &  35 & 5 & 15 & 25 & 35  \\
        \midrule
         \multirow{2}{*}{WorkBench} & DxD &   $\textbf{41.53}$ & $39.24$ & $38.03$ & $40.25$  &  $\textbf{50.65}$ & $34.15$ & $35.85$ & $40.09$ & $\textbf{17.76}$ & $16.53$ & $14.90$ & $15.51$ 
 \\
         & GD &   $28.86$ & $30.07$ & $28.68$ & $24.74$  &  $32.37$ & $21.96$ & $22.93$ & $15.46$  & $15.01$ & $15.00$ & $14.60$ & $14.81$  \\
        \multirow{2}{*}{(55 Tasks)} & GDEC &  $32.56$ & $26.36$ & $26.30$ & $27.52$  &  $40.37$ & $28.87$ & $26.09$ &  $30.44$ & $15.97$ & $15.16$ & $14.76$ & $15.37$  \\
        &  OD &   \multicolumn{4}{c|}{$24.16$}  &  \multicolumn{4}{c|}{$17.19$}  & \multicolumn{4}{c}{$17.31$}  \\
        \midrule
             \multirow{2}{*}{$\tau$-Bench }  & DxD & $48.49$ & $12.89$ & $0.00$ & $0.00$  & $42.12$  & $0.00$ & $0.00$ & $0.00$  &  $54.59$ & $0.00$ & $0.00$ & $0.00$  \\
         & GD &  $10.44$ & $5.62$ & $6.43$ & $8.03$  & $35.34$ & $27.31$ & $38.15$ & $21.29$ & $27.71$ & $31.33$ & $21.69$ & $13.25$  \\
        (77.6 Tasks)& GDEC &  $40.96$ & $40.16$ & $34.94$ & $34.94$ &$56.63$ & $66.27$ & $65.06$ & $51.0$ & $58.23$   & $53.82$ & $53.01$ & $39.76$  \\
        &  OD &  \multicolumn{4}{c|}{$\textbf{78.59}$}   &  \multicolumn{4}{c|}{$\textbf{69.08}$}   &    \multicolumn{4}{c}{$\textbf{86.30}$}  \\
        \midrule
             \multirow{2}{*}{CRMArena}  & DxD & $69.24$ & $67.22$ & $0.00$ & $0.00$  & $12.02$ & $0.00$ & $0.00$ & $0.00$  & $66.33$ & $0.00$ & $0.00$ & $0.00$  \\
         & GD &  $52.30$ & $49.86$ & $18.86$ & $39.22$  &$0.0$ & $0.0$ & $0.0$ & $0.0$  &  $8.20$ & $14.19$ & $15.84$ & $14.38$   \\
        \multirow{2}{*}{(183 Tasks)}& GDEC &  $60.49$ & $51.88$ & $45.92$ & $9.96$ & $7.51$ & $11.52$ & $0.0$ & $0.0$ & $54.13$ & $49.91$ & $0.00$ & $0.00$      \\
        &  OD &  \multicolumn{4}{c|}{$\textbf{72.14}$}  &   \multicolumn{4}{c|}{$\textbf{29.26}$} &  \multicolumn{4}{c}{$\textbf{84.53}$}  \\
        \bottomrule

    \end{tabular}}
    \caption{The models specified in the top row indicate the LLM used as the API-based agent and the document generator. All experiments ran across $3$ trials, and all values reported are the mean success rate (\%). Bolded values are the highest SR given a dataset and model.}
    \label{tab:main_results}
\end{table*}

\section{Results: Learning Parameter Information Is Crucial}
From our experiments, a recurring problem at every stage of the pipeline (document generation, task completion, and evaluation) is the parameter filling. Parameter filling has also been indicated as a problem by \citet{stylesworkbench, yao2024tau} for WorkBench and $\tau$-Bench, respectively. We now show how the problem increases in difficulty in scenarios with no ground-truth documentation.
\noindent \textbf{Document Generation Has Problems With Specifying Parameter Information.} One main issue is generating parameter information. Many times, generators hallucinated input parameters that did not exist or left out input parameters seen in the demonstrations. This occurred even when in the system prompt we stated not to hallucinate and only focus on the parameters it saw. Thus, given a set of demonstrations for $f$, we programmatically found all used input parameters and explicitly mentioned in the user prompt to generate information on only these parameters. That solved the issue for the OpenAI models and Gemma. However, documents generated by Mistral many times do not include parameter information (see Figure \ref{fig:mistral_analytics} in the Appendix). Other sources of error were more specific, such as when generated documents would specify to include timezone information for time\_min and time\_max parameters in WorkBench functions. Agents that followed this got a runtime error. The generated documents from more demonstrations would more frequently make this mistake, and also say that the time\_min and time\_max parameters were required when they were optional. Documents generated from fewer demonstrations would, in some trials, correctly say that those parameters were optional, allowing the agent to not fill in those parameters with incorrect formatting. We hypothesize that the more repeated use of these parameters in the larger set of demonstrations causes the document generator to believe that those are required, causing the degradation in Table \ref{tab:main_results} as $N$ increases.

\noindent \textbf{Generated Guidelines Can Help Parameter Understanding.} Attaching generated guidelines (AG) has a $12\%$ improvement over (GD) in WorkBench. In a representative case, ``Kerry Moore is no longer a customer. Can you delete them from the crm?'', the GD agent tries to directly delete the customer by using their name as the customer\_id parameter, rather than searching for their ID and then deleting. The guidelines for customer.delete\_customer state to avoid directly deleting customers and ensure that the customer ID is a valid value, not someone's name. This result shows that self-exploration with an evaluation protocol can enable self-improvement.

\begin{figure*}
    \centering
    \includegraphics[width=\linewidth]{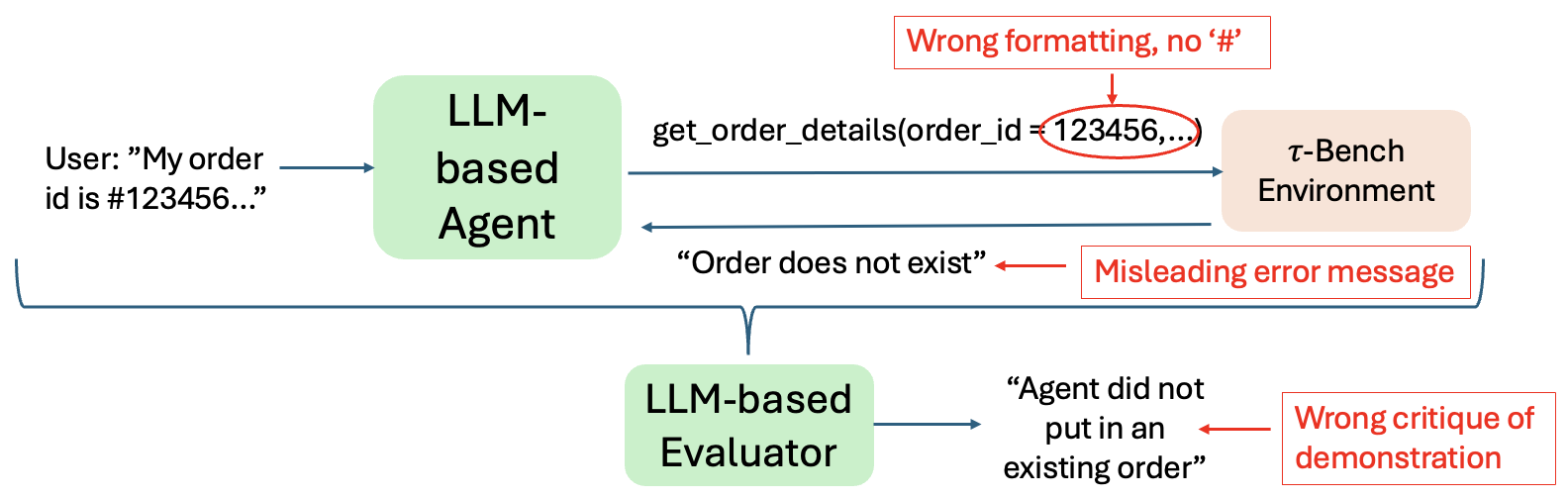}
    \caption{\textbf{Failure of System Due to Misleading Error Messages.} Incorrectly formatting the order\_id parameter value for an order that does exist causes the $\tau$-Bench function to return ``Order does not exist'', misleading the LLM-based agent and LLM-based evaluator.}
    \label{fig:error_misleading}
\end{figure*}

\noindent \textbf{Error Handling of Parameter Formatting Affects Performance.} However, we see that for $\tau$-Bench, AG, and all other methods except for OD, achieved a low score with $\tau$-Bench. A major contributing factor is the order\_id parameter. The order\_id parameter is in $7$ out of $14$ functions in $\tau$-Bench, and requires a `\#' at the beginning of the value; otherwise, an error is returned. When the order\_id is formatted incorrectly, the error message ``Order not found'', shown in Figure \ref{fig:error_misleading}, is misleading to the agent and the self-improvement system because there is no indication that the missing ``\#'' is the problem of the error. Some of the generated evaluations marked some of the function calls that used the `\#' as wrong. This error in evaluation then affects the summarized guidelines. In $2$ trials, the guidelines would say to remove the `\#' formatting from IDs. In $\tau$-Bench, none of the generated documentation methods (GD, UD, and RD) mention the formatting requirement either. Therefore, the only way to know that the ``\#'' is required was by using the OD. We reran the AG (with DxD method before self-exploration) experiment. However, this time we hardcoded the following phrase in all the guidelines for all functions that use order\_id: ``THERE NEEDS TO BE A `\#' AT THE START OF THE ORDER ID. THIS IS MANDATORY. FOR EXAMPLE: ``\#W8732376'' IS VALID, BUT ``W8732376'' IS NOT!!!!'' This modification raises the average success rate from $10\%$ to $49\%$, and this result shows again that having specific, correct parameter information can significantly help the SR, and that robust and descriptive error handling of the API functions is imperative for reliable evaluation and summarization.

\noindent \textbf{DxD and DE have more advantages than the summarization methods.} DxD and DE have consistently had a higher success rate over other $I$ and $I'$ methods. However, this can be attributed to data leakage. The set of tasks $G$ for each environment has many repeated formats, such as sending emails, canceling orders, or looking up information on cases. Thus, the train-test distributions are similar. It would be interesting to try DxD and DE in environments that provide distinctive tasks that use the same APIs.



\noindent \textbf{Note on Increasing $N$.} For DxD with $\tau$-Bench and CRMArena, using the conversation history between causes the LLM of the agent to exceed its maximum context length. We truncated the conversation history to at most the past three messages. This truncation allowed o3-mini to handle $5$ and $15$ demonstrations per function for $\tau$-Bench, and $5$ per function for CRMArena. All other models could only handle $5$ demonstrations per function. For GDEC, we noticed that for WorkBench and CRMArena, the agent will repeatedly call the same function, suggesting that the repeated $N$ example calls teach the LLM agent to repeat.

\begin{table}
    \centering
    \small
    \resizebox{0.9\columnwidth}{!}{\begin{tabular}{c|cc|cc}

            Test-Time  & \multicolumn{2}{c|}{WorkBench (167 Tasks)}  &  \multicolumn{2}{c}{$\tau$-Bench (83 Tasks)}  \\
            \cmidrule{2-5} 
                 Method  & \multicolumn{4}{c}{$I$ Method Before Self-Exploration}     \\
                 \cmidrule{2-5}
                 & DxD & GD & DxD & GD  
                 \\
                 \midrule
                DE & $\textbf{44.71}$&  $32.14$  & $45.00$ & $44.00$    \\
              UD  & $29.20$ & $28.94$  & $9.00$ & $9.00$ \\
              RD   & $30.34$  &  $28.54$   & $6.00$  &  $10.00$  \\
             AG   & $42.51$  &  $38.92$ & $10.00$ &  $9.00$  \\
                \midrule
                 DxD   &\multicolumn{2}{c|}{ $44.11$ }   &  \multicolumn{2}{c}{$47.00$}   \\
            GD   & \multicolumn{2}{c|}{ $30.14$}  &  \multicolumn{2}{c}{$9.00$}  \\
              GDEC  & \multicolumn{2}{c|}{ $34.73$}  &   \multicolumn{2}{c}{$41.00$}  \\
         OD & \multicolumn{2}{c|}{ $33.33$} & \multicolumn{2}{c}{$\textbf{78.00}$}  \\
    \bottomrule
    \end{tabular}}
    \caption{ Results of gaining experience with training queries during self-exploration using o3-mini. We used $N=5$ expert demonstrations per $f$ for self-exploration. The methods on the bottom half of the table (DxD, GD, GDEC, OD) do not depend on self-exploration. All experiments ran across $3$ trials, and all values reported are the mean success rate (\%). Bolded values represent the highest success rate for a given dataset.}
    \label{tab:main_results2}
\end{table}

\section{Additional Analysis and Results}

\textbf{Retrieval-Augmented Generation (RAG).} To handle the context length issue seen with GPT-4o-mini, we provide an additional RAG \cite{lewis2020retrieval} experiment on $\tau$-Bench using DxD. We embed the sampled N demonstrations with OpenAI text-embedding-3-small and retrieve the $5$ most relevant demonstrations based on the current query with cosine similarity. This experiment is over $3$ trials with the set of questions that use APIs with at least $35$ demonstrations, same as in Table \ref{tab:main_results}. Table \ref{tab:rag} shows no significant change in performance in number of demonstrations when using RAG.

\begin{table}[h]
\centering
\small
\resizebox{\columnwidth}{!}{\begin{tabular}{lcccc}
\textbf{OD} & \textbf{5 Dems } & \textbf{15 Dems} & \textbf{35 Dems} \\
(No RAG)& (No RAG) & & 
\\
\midrule
$69.08 $ & $42.12$ & $39.13$ & $44.26 2$ \\
\bottomrule
\end{tabular}}
\caption{Mean success rate (\%) across different demonstration counts with RAG on $\tau$-Bench.}\label{tab:rag}
\end{table}

\noindent \textbf{DxD with Open-Source Models.} We ran Mistral 8x7B and Qwen3 on three trials of WorkBench with OD, and DxD (with 5 Dems). The test tasks are the same set of questions from Table \ref{tab:main_results}, where APIs needed have at least 35 demonstrations. In Table \ref{tab:open_source}, Qwen3 achieves higher SR with DxD than with OD, and it outperforms all OpenAI models on WorkBench. We believe that this increase is due to the amount of prior reasoning it does before generating an action, and the similarity between the train and test split in WorkBench.

\begin{table}[h]
\centering
\begin{tabular}{lcc}
\textbf{I Method} & \textbf{Qwen3} & \textbf{Mistral 8x7B} \\
\midrule
DxD 5 Dems & $67.27 $ & $13.28 $ \\
OD         & $54.62$ & $14.56 $ \\
\bottomrule
\end{tabular}
\caption{Comparing 5 DxD against OD with Qwen-3 and Mistral 8x7B on WorkBench.}\label{tab:open_source}
\end{table}


\section{Related Works}

Tool-based agents expand the capabilities of traditional LLMs by connecting to external sources such as API functions, search engines, data centers, etc., to complete multi-step queries. Multiple benchmarks have been proposed with various sets of API functions with documentation \citep{guo2024stabletoolbench, liu2024apigen, huang-etal-2025-crmarena, stylesworkbench, yao2024tau, xu2024theagentcompanybenchmarkingllmagents, arcadinho2024automated}. While generating documentation has been studied before the advent of LLMs \cite{wang2023gdoc, nybom2018systematic}, there is little work on generating documentation for downstream agent task planning. For documentation-free agents, ToolAlpaca by \citet{tang2023toolalpaca} generated documentation via an LLM given a brief description of the function from OpenAPI. APIDocBooster \cite{yang2023apidocbooster} used GPT-4 to produce and update documentation based on search results from StackOverflow. Concurrent to this work, \citet{fang2025synworldvirtualscenariosynthesis} also used experiences of the agent to update the documentation. However, they still relied on initial documentation. Toolformer by \citet{schick2023toolformer} finetuned their agent on API calls. Our work is the first to study how we can generate \textit{and} update an agent's understanding of functionality with expert demonstrations and experience with a frozen LLM. We are also the first to use demonstrations directly into the LLM-based agent for functionality understanding. Our formalization is similar to Reinforcement Learning with parameterized action spaces \citep{masson2016reinforcement, zhang2024model} where the agent must choose an action from a discrete set (in our case, choose an $f \in \mathcal F$), and then must choose parameters specific to the selected action, i.e. $p \in \text{dom}(P(f))$. However, to our knowledge, previous literature assumes a single, continuous parameter per action. Our work relaxes this assumption as an API function can allow for multiple parameters, continuous or discrete.

\section{Conclusion}

We provide and formalize the problem of earning of API functionality from in-context demonstrations with no prior documentation. We highlight the persisting and challenging problem of API-agent planning with limited information on the set of functions. We investigate tackling this problem by learning functionality, specifically function description, return, and input parameters, from in-context demonstrations of tool calls. We analyze the number of demonstrations, various processing methods, and the impact of self-exploration and LLM-based evaluation. Importantly, our extensive experiments highlight the difficulties of SoTA LLMs on this problem, specifically due to failures in describing the parameter schema, suggesting further research to improve results. 

\section{Limitations and Further Work}

As we present this new challenge, we would like to highlight some limitations of this work. The heuristic mentioned in Section \ref{sec:experimental_setup} when extracting expert demonstrations leads to suboptimal demonstrations. We do not focus on mitigating or filtering out these suboptimal demonstrations. One could also look at using an LLM-based evaluator to signal what steps are correct. We do not consider it in this work, as we analyzed LLM-based evaluations during self-exploration. However, filtering out suboptimal demonstrations is an exciting and interesting direction to pursue. 



Furthermore, we iteratively generated documents for each function independently. One interesting extension is to see the effect of group learning functions together. We see in $\tau$-Bench that the order\_id parameter was shared across multiple functions, so the agent should only have to learn it correctly once. This sharing of parameter information could lead to more stable performance as the information between functions is more consistent.

One could also extend this problem to the Model-Context Protocol (MCP), where the agent must rely on multiple data sources and sets of API functions. This direction is a more complex, realistic scenario than ours, as we only focus on a single set of data and APIs. Furthermore, human-in-the-loop setups where the user acts as an expert policy to extract online demonstrations could pave the way to incorporate imitation learning algorithms, such as DAgger \citep{ross2011reduction}.

\bibliography{main}
\bibliographystyle{Styles/acl/acl_natbib}

\appendix
\onecolumn

\section{Train-Test Split}\label{sec:train_test_appendix}

Algorithm \ref{alg:train-test} details the process of how, for a given dataset, we divide the demonstrations of the API functions into the train and test sets. Because a task in a dataset could require different APIs, had we simply just split the demonstrations, which include task information, into train and test sets, data leakage would almost certainly occur. To ensure no tasks in the test sets were in the training sets, for each API function, we first divide the tasks from only the demonstrations of that function into train and test task sets. We then divide the demonstrations into training and test demonstration sets. For the next API function, if a task for that API has already been assigned to the train or test task set, we simply assign the demonstration to the corresponding set for demonstrations. The remaining tasks for the function that have not been seen will be split into the train and test task sets, and the process repeats for all functions.

\begin{algorithm}[!h] 
\caption{Train-Test Split}\label{alg:train-test}
\begin{algorithmic}
\State \textbf{Input:} List of API functions $A$, dictionary of demonstrations $D$, test split percentage $p$, random seed $s$
\State \textbf{Initialize:} List of training tasks $T_{train}$ = []; list of test tasks $T_{train}$ Dictionary of training demonstrations $D_{train} = {}$, Dictionary of training demonstrations $D_{test} = {}$

\For{$a$ in $A$} 
\State $T_{remaining}^a = \{\}$ \Comment{See which tasks of $a$ have not been assigned to either train or test}
\For {$d$ in $D[a]$} \Comment{$D[a]$ is the list of demonstrations for $a$}
\State $t = d[{Task}]$ \Comment{Get the task that the demonstration is for} 
\If{$t$ in $T_{train}$} 
    \State Add $d$ to list $D_{train}[a]$ 
\ElsIf{$t$ in $T_{test}$}
    \State Add $d$ to list $D_{test}[a]$ 
\Else                   
    \State Add $t$ to list $T_{remaining}Ya$ 
\EndIf
\EndFor
\State Set random state to seed $s$ \Comment{We now split the tasks that have not been assigned yet}
\State $num_{test} = len(T_{remaining}^a) * p$ 
\State $T_{test}^a =$ random $n$ subset of $T_{remaining}^a$
\State $T_{train}^a = T_{remaining}^a - T_{test}^a$  
\For {$d$ in $D[a]$}
\State $t = d[{Task}]$
\If{$t$ in $T_{train}^a$}
\State Add $d$ to $D_{train}[a]$
\Else
\State Add $d$ to $D_{test}[a]$
\EndIf
\EndFor
\State Add elements of $T_{train}^a$ to $T_{train}$ 
\State Add elements of $T_{test}^a$ to $T_{test}$
\EndFor
\end{algorithmic}
\end{algorithm}

\section{AI Writing Tools}

We used LLM services such as ChatGPT and Perplexity for feedback and editing of this document.

\section{Dataset Statistics and  Specifications}\label{sec:dataset_appendix}

\textbf{WorkBench.} To extract the expert demonstrations, we take their pre-computed predictions for all queries and then run their evaluation script. From there, we have access to the prediction of their GPT-4 agent that used the original documentation. The prediction is used as the demonstration trajectory. For each function call of prediction, we format a demonstration as shown in Figure \ref{fig:dem_example}. Because WorkBench is outcome-centric, the ground-truth will contain fewer steps than needed to properly complete the task. The only functions that are included affect the environment. For example, sending an email to someone, a task that involves finding the email address, will only have the search\_email function in the ground-truth. All functions that search for events, customers, emails, etc., or analyze data like analytics.get\_average\_session will not appear. Some queries will have ground truths with $0$ steps. These are for queries where the required condition before performing an environment-changing action is not satisfied. For example, the agent needs to set up a meeting only if the last meeting was over a week ago, and the last meeting was the day before. Then, the ground truth is $0$ steps. However, the agent should execute the search\_events function before deciding not to set up an email. For this reason, we use the pre-computed trajectories provided by \citep{stylesworkbench} with a GPT-4 powered agent with ground-truth documentation. We only extract demonstrations from successful trajectories. This heuristic, as previously mentioned in Section \ref{sec:experimental_setup}, can lead to wrong demonstrations in the set of expert demonstrations. Table \ref{tab:workbench_dem_stats} provides the number of expert demonstrations we extracted from the pre-computed trajectory.

\begin{table}
    \centering
    \begin{tabular}{c|c}
    Function from WorkBench & No. of Demonstrations \\
    \midrule
    analytics.create\_plot & $152$ \\
    analytics.engaged\_users\_count & $46$ \\  
    analytics.get\_average\_session\_duration & $33$ \\  
    analytics.total\_visits\_count & $28$ \\  
    analytics.traffic\_source\_count & $2$ \\  
    calendar.create\_event & $93$ \\  
    calendar.delete\_event & $118$ \\  
    calendar.search\_events & $85$ \\  
    calendar.update\_event & $52$ \\  
    company\_directory.find\_email\_address & $67$ \\  
    customer\_relationship\_manager.add\_customer & $25$ \\  
    customer\_relationship\_manager.delete\_customer & $175$ \\  
    customer\_relationship\_manager.search\_customers & $13$ \\  
    customer\_relationship\_manager.update\_customer & $133$ \\  
    email.delete\_email & $44$ \\  
    email.forward\_email & $67$ \\  
    email.reply\_email & $31$ \\  
    email.search\_emails & $62$ \\  
    email.send\_email & $108$ \\  
    project\_management.create\_task & $78$ \\  
    project\_management.search\_tasks & $34$ \\  
    project\_management.update\_task & $96$ \\  

    \end{tabular}
    \caption{WorkBench functions and Number of Expert Demonstrations}
    \label{tab:workbench_dem_stats}
\end{table}

\begin{figure}
    \centering
    \includegraphics[width=0.75\linewidth]{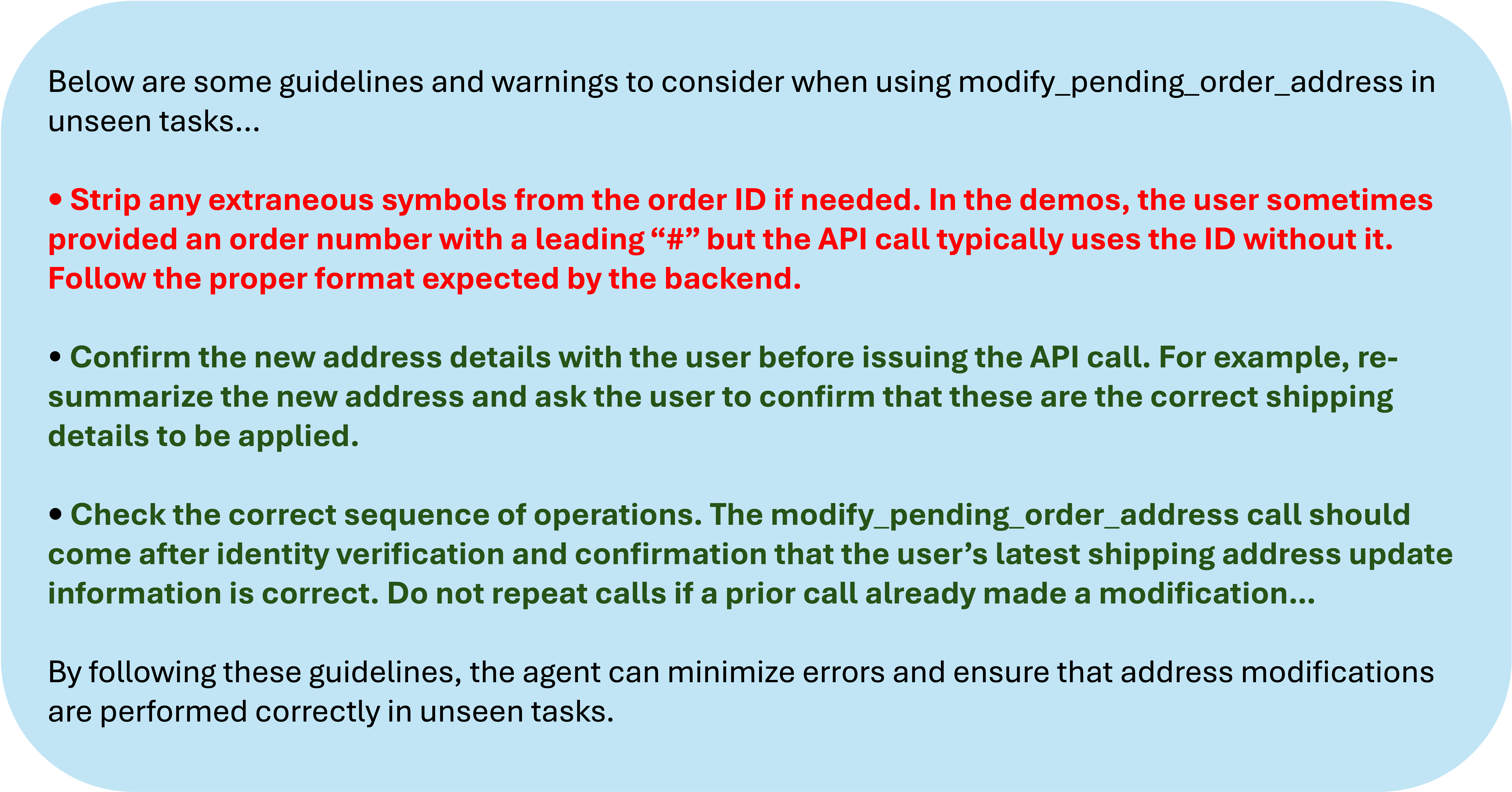}
    \caption{Generated guidelines for the modify\_pending\_order\_items function in $\tau$-Bench}
    \label{fig:modify_pending_guidelines}
\end{figure}

\noindent \textbf{$\tau$-Bench and CRMArena.} Both $\tau$-Bench and CRMArena have their results stored as JSON objects. Each task attempted stores a sequence of tool calls and accompanying tool executions. Each pair of tool calls and tool executions is turned into the demonstration data, where we can also extract the Return value from the tool execution to add to the expert demonstrations. While $\tau$-Bench does provide ground-truth function calls for the tasks, it disregards the interaction with the simulated user. Seeing examples of these interactions is important for the agent to understand when to use tools, relating to parameter filling. We thus do not use the ground-truth answers and rather use the pre-computed trajectories from the dataset as expert demonstrations. Figure \ref{fig:modify_pending_guidelines} shows generated guidelines that specify, in green, that the agent should confirm the input parameters of the address details with the user.

\begin{table}
    \centering
    \begin{tabular}{c|c}
    Function from $\tau$-Bench & No. of Demonstrations \\
    \midrule
    calculate & $36$ \\  
    cancel\_pending\_order & $129$ \\  
    exchange\_delivered\_order\_items & $133$ \\  
    find\_user\_id\_by\_email & $245$ \\  
    find\_user\_id\_by\_name\_zip & $320$ \\  
    get\_order\_details & $871$ \\  
    get\_product\_details & $340$ \\  
    get\_user\_details & $279$ \\  
    list\_all\_product\_types & $74$ \\  
    modify\_pending\_order\_address & $43$ \\  
    modify\_pending\_order\_items & $146$ \\  
    modify\_pending\_order\_payment & $12$ \\  
    modify\_user\_address & $16$ \\  
    return\_delivered\_order\_items & $140$ \\  
    \end{tabular}
    \caption{$\tau$-Bench functions and Number of Expert Demonstrations}
    \label{tab:taubench_dem_stats}
\end{table}

\begin{table}
    \centering
    \begin{tabular}{c|c}
    Function from CRMArena & No. of Demonstrations \\
    \midrule

    calculate\_average\_handle\_time & $62$ \\  
    calculate\_region\_average\_closure\_times & $114$ \\  
    find\_id\_with\_max\_value & $175$ \\  
    find\_id\_with\_min\_value & $144$ \\  
    get\_account\_id\_by\_contact\_id & $63$ \\  
    get\_agent\_handled\_cases\_by\_period & $136$ \\  
    get\_agent\_transferred\_cases\_by\_period & $97$ \\  
    get\_agents\_with\_max\_cases & $113$ \\  
    get\_agents\_with\_min\_cases & $32$ \\  
    get\_cases & $421$ \\  
    get\_email\_messages\_by\_case\_id & $53$ \\  
    get\_issue\_counts & $166$ \\  
    get\_issues & $83$ \\  
    get\_livechat\_transcript\_by\_case\_id & $53$ \\  
    get\_month\_to\_case\_count & $38$ \\  
    get\_non\_transferred\_case\_ids & $59$ \\  
    get\_order\_item\_ids\_by\_product & $304$ \\  
    get\_period & $290$ \\  
    get\_purchase\_history & $172$ \\  
    get\_qualified\_agent\_ids\_by\_case\_count & $218$ \\  
    get\_shipping\_state & $82$ \\  
    get\_start\_date & $240$ \\  
    search\_knowledge\_articles & $114$ \\  
    search\_products & $195$ \\  
    \end{tabular}
    \caption{CRMArena functions and Number of Expert Demonstrations}
    \label{tab:crm_dem_stats}
\end{table}

\section{Self-Exploration Pipeline}

\begin{figure*}[!h]
    \centering
    \includegraphics[width=0.8\textwidth]{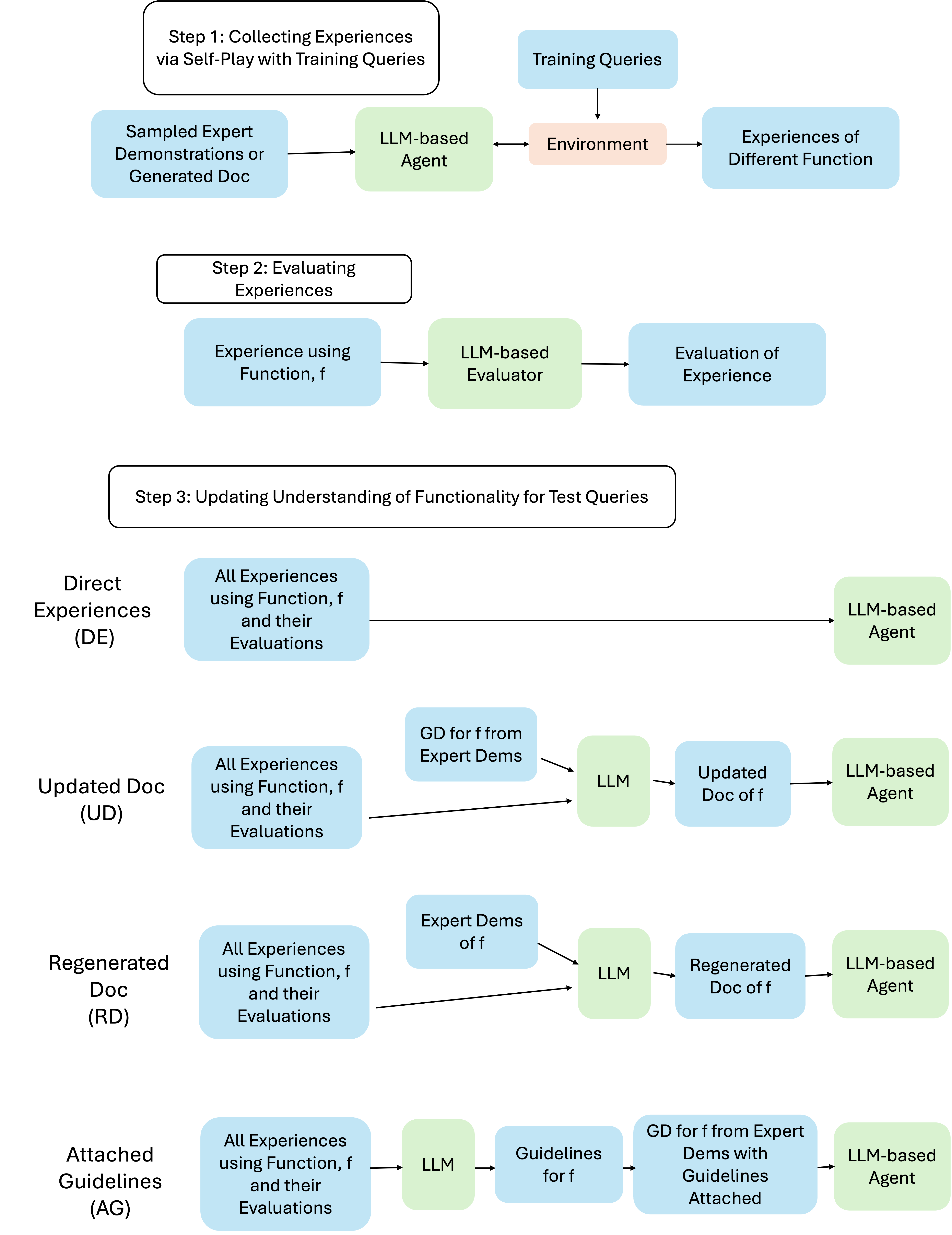}
    \caption{\textbf{Overview of Self-Exploration Pipeline.} Visualizes is the three-step process of updating functionality understanding with experience. In the first step, the agent performs self-exploration with the training queries in the sandbox environment, collecting experiences. The agent can use any method $I$ to initially learn functionality from expert demonstrations. In the second step, each experience is then evaluated via an  LLM-based evaluator. Finally, we present four different options to update the agent's understanding of functionality with the original expert demonstrations and the new experience-based demonstrations. The updated understanding is then passed into the agent for test queries.}
    \label{fig:self_play_overview}
\end{figure*}

Figure \ref{fig:self_play_overview} visualizes the process of updating the agent's initial learning from in-context demonstrations via self-exploration.

\newpage

\section{System Prompts}\label{sec:prompt_appendix}

\begin{figure*}[!h]
    \centering
    \includegraphics[width=\linewidth]{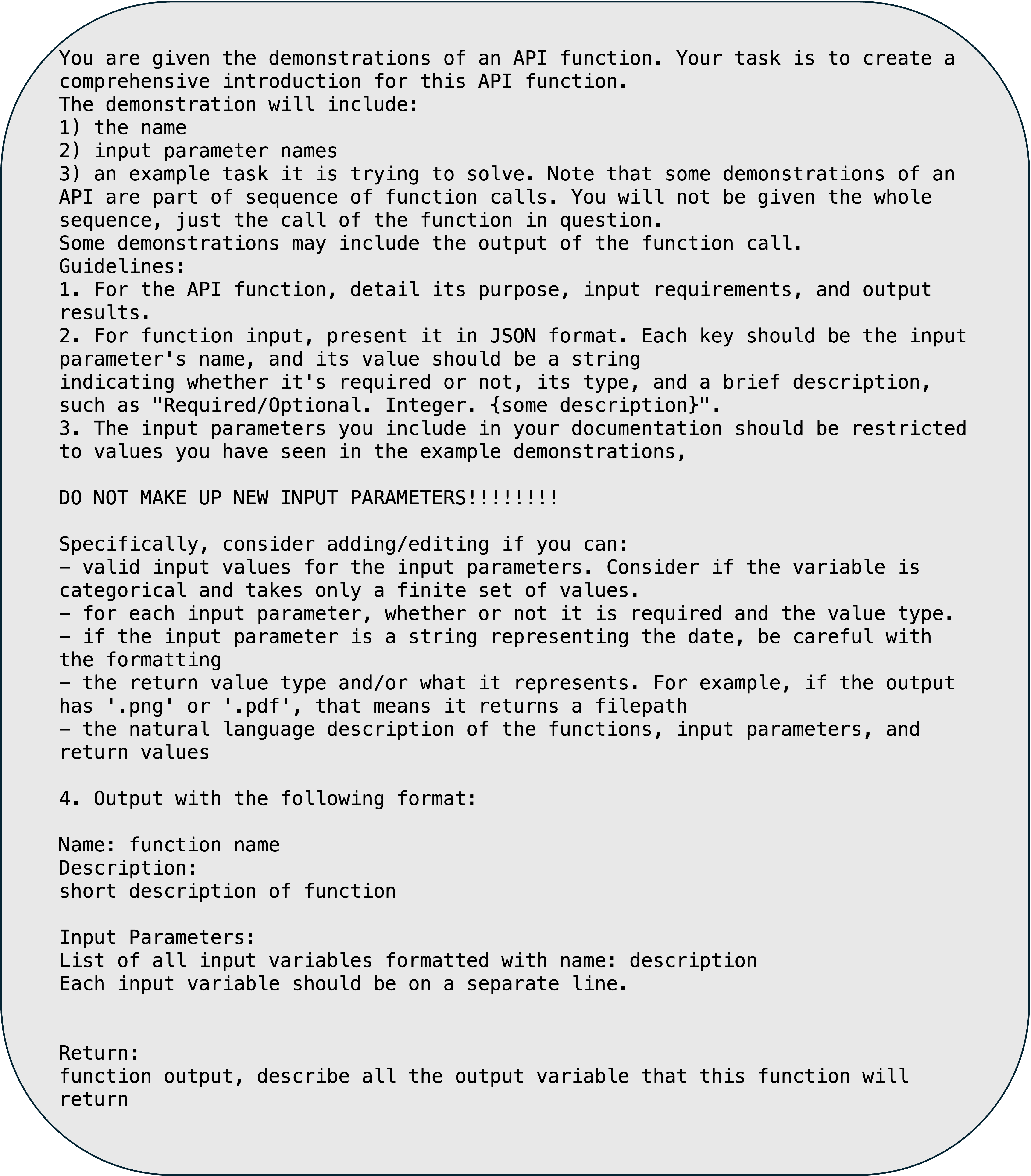}
    \caption{System Prompt of document generator }
    \label{fig:doc_gen_prompt}
\end{figure*}

\begin{figure*}[!h]
    \centering
    \includegraphics[width=\linewidth]{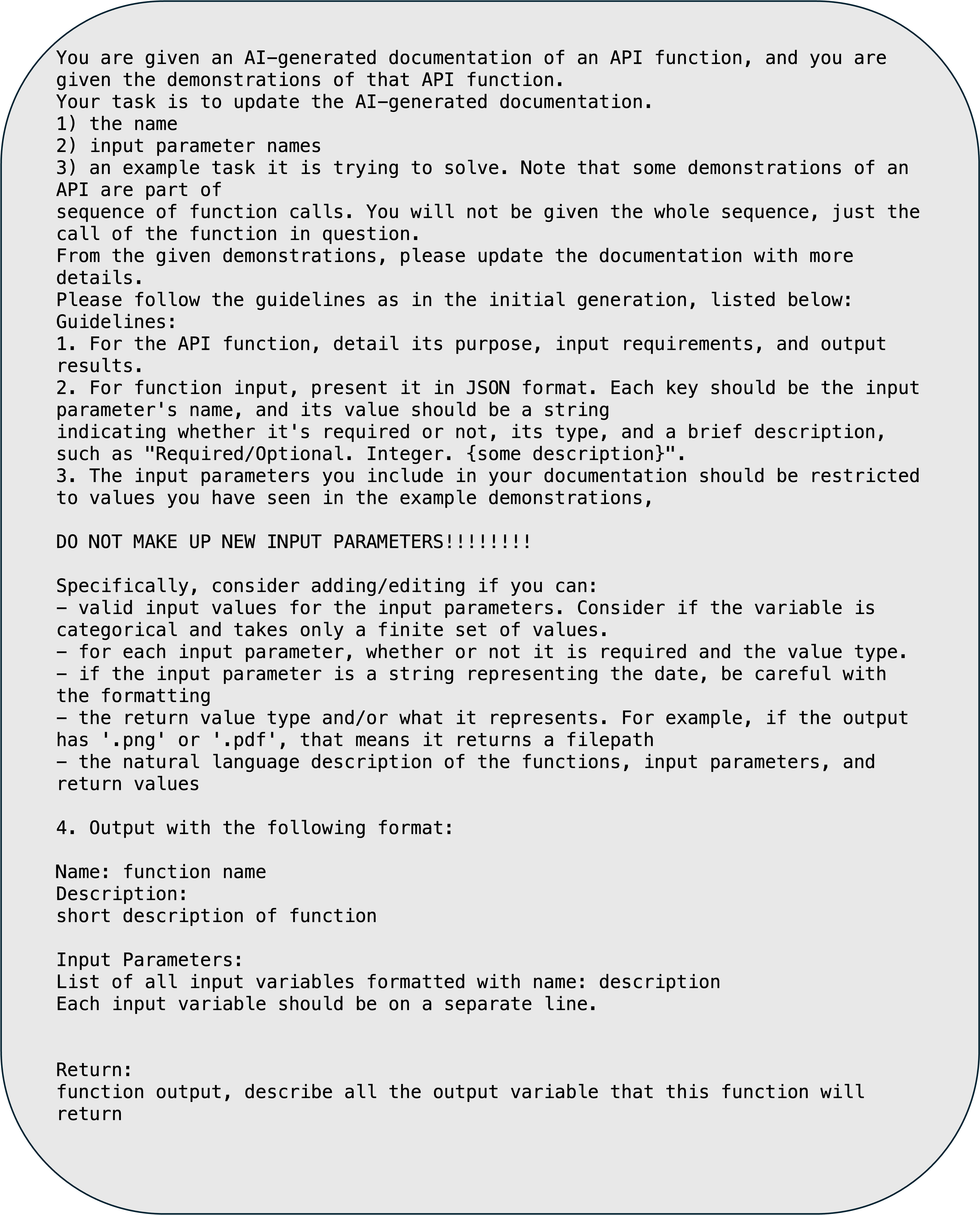}
    \caption{System Prompt of documentation updater given initial generated document and experiences}
    \label{fig:doc_update_prompt}
\end{figure*}




\begin{figure*}[!h]
    \centering
    \includegraphics[width=\linewidth]{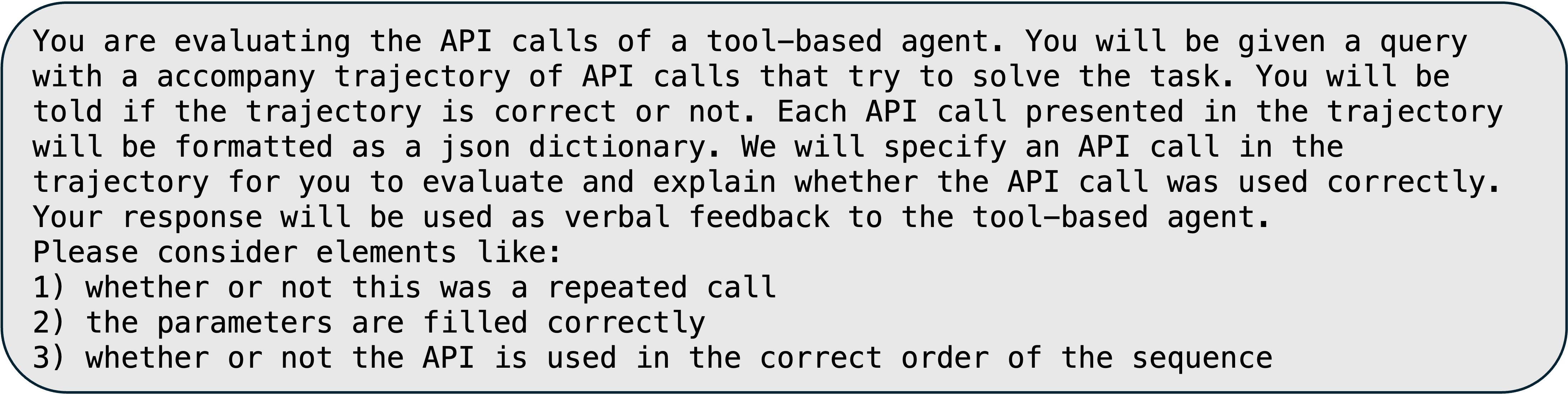}
    \caption{System Prompt of evaluator that critiques an experience during self-exploration}
    \label{fig:online_dem_eval_prompt}
\end{figure*}

\begin{figure*}[!h]
    \centering
    \includegraphics[width=\linewidth]{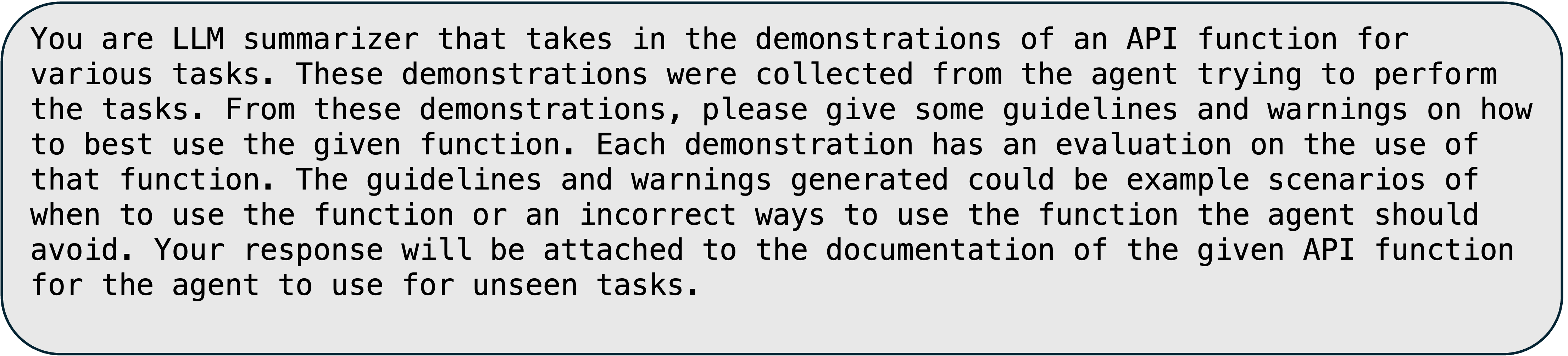}
    \caption{System Prompt that summarizes experiences and evaluations to generate guidelines}
    \label{fig:doc_summarizer_prompt}
\end{figure*}


\begin{figure}
    \centering
    \includegraphics[width=0.75\linewidth]{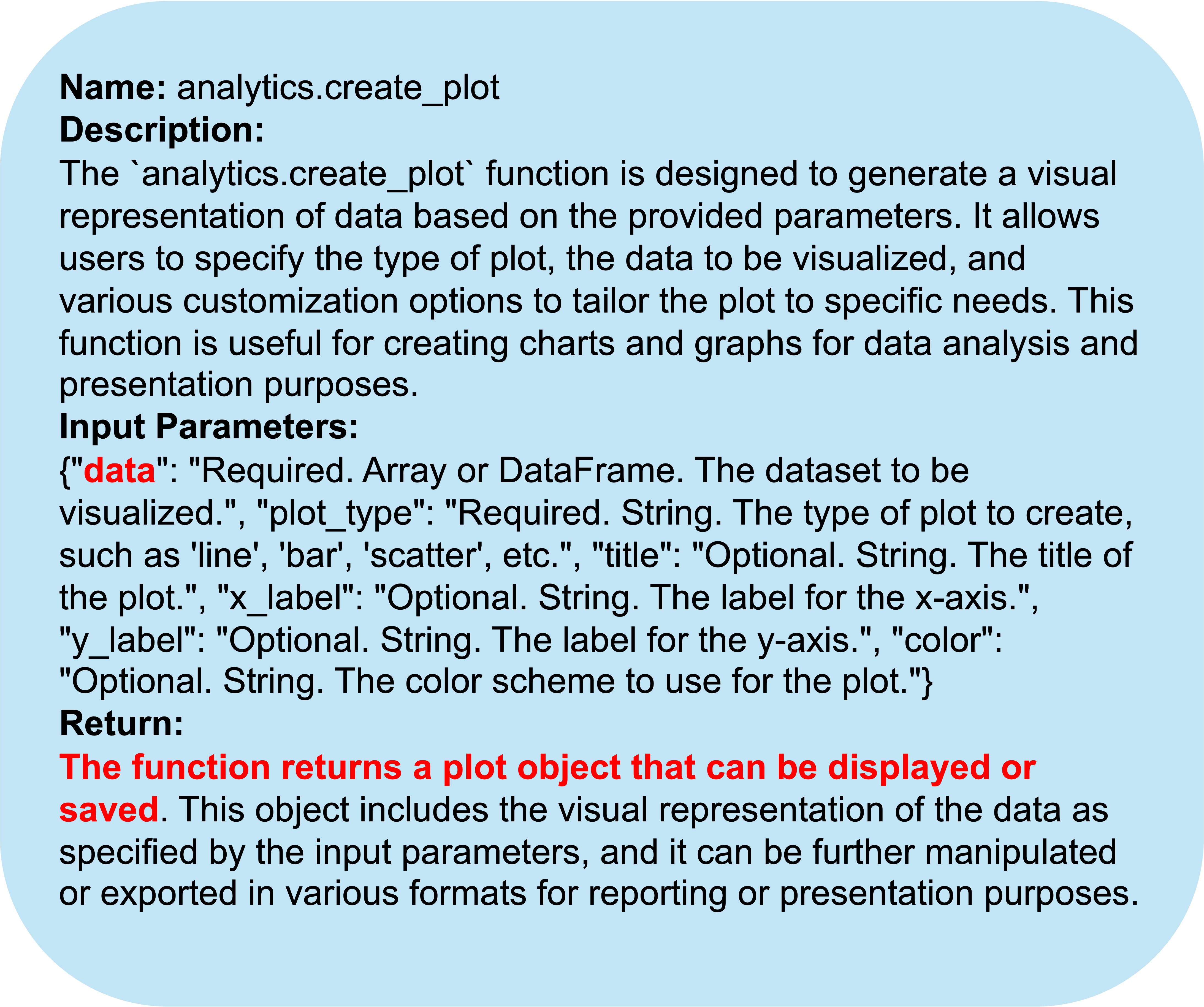}
    \caption{GPT-4o Generated Document of analytics.create\_plot with 10 demonstrations before we explicitly stated in the user prompt what the input parameters should be. GPT-4o hallucinates a `data' input parameter that does not exist in the implementation. }
    \label{fig:gpt4o_hallucination}
\end{figure}

\begin{figure}
    \centering
    \includegraphics[width=0.75\linewidth]{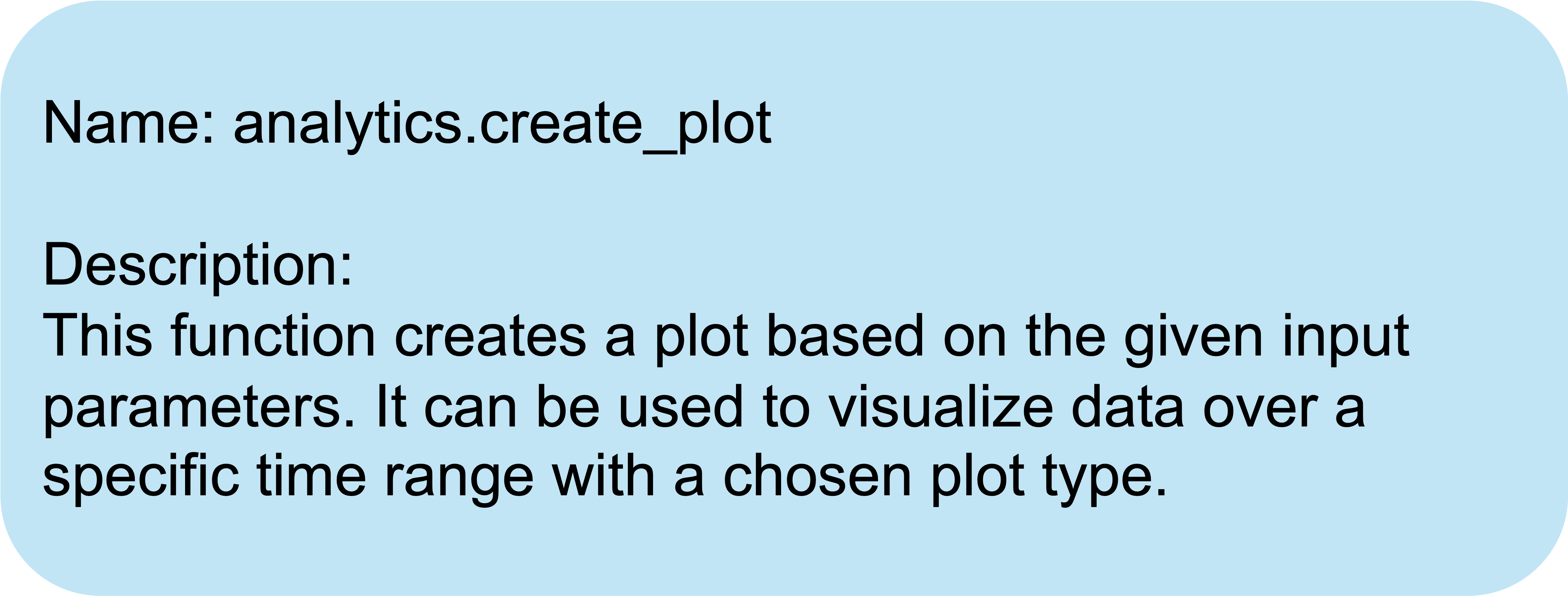}
    \caption{Mistral Generated Document of analytics.create\_plot with 5 demonstrations. It has no parameter information.}
    \label{fig:mistral_analytics}
\end{figure}

\end{document}